\documentclass[conference]{IEEEtran}
\usepackage{cite}
\usepackage{amsmath,amssymb,amsfonts}
\usepackage{graphicx}
\usepackage{textcomp}
\usepackage{xcolor}
\usepackage{tikz}
\usepackage{algorithm}
\usepackage{algpseudocode}
\usepackage{float}
\usetikzlibrary{arrows,shapes,trees}
\def\BibTeX{{\rm B\kern-.05em{\sc i\kern-.025em b}\kern-.08em
    T\kern-.1667em\lower.7ex\hbox{E}\kern-.125emX}}

\title{An Enhanced RRT based Algorithm for Dynamic Path Planning and Energy Management of a Mobile Robot
}

\author{Ronit Chitre and Arpita Sinha}
\thanks{Ronit Chitre is with Aerospace Engineering, Indian Institute of Technology, Bombay, Mumbai, India, and Arpita Sinha is with Systems and Control Engineering, Indian Institute of Technology Bombay, Mumbai, India
{\tt \small{chitreronit@gmail.com, arpita.sinha@iitb.ac.in}}
}

\begin{document}

\maketitle
\begin{abstract}
Mobile robots often have limited battery life and need to recharge periodically. This paper presents an RRT-based path-planning algorithm that addresses battery power management. A path is generated continuously from the robot's current position to its recharging station. The robot decides if a recharge is needed based on the energy required to travel on that path and the robot's current power. RRT* is used to generate the first path, and then subsequent paths are made using information from previous trees. Finally, the presented algorithm was compared with Extended Rate Random Tree (ERRT) algorithm \cite{c4}. 

\end{abstract}

\begin{IEEEkeywords}
RRT, path planning, robot energy management, autonomous systems
\end{IEEEkeywords}
\section{Introduction} \label{intro}
This paper addresses a path planning problem with a battery power management system. Not all tasks are doable with one battery charge. We may need to recharge the battery several times before completing the task. An example can be a battery-powered autonomous robot harvesting a huge field. The robot must know when to travel to a charging point in such cases. A conservative approach will increase the total time to complete the job. The other extreme may lead to the robot running out of battery before it reaches the recharge point. In this paper, we develop an online algorithm that indicates the best time for the robot to return for recharging. 

We use RRT* to plan the path from the current robot location to the recharge station to find the appropriate time to return to the base. However, the robot follows a path to complete its task and the return-to-base path is taken only when required. Since the return-to-base path planning needs to happen every instant, RRT* generates a new tree every time. We propose to use the trees built previously to reduce the computational time of RRT*. This concept is similar to the RRT algorithms applied to a dynamic environment. However, unlike the dynamic environment scenario where the robot moves on the RRT algorithm's path, here the robot follows a track to execute the task independently of the RRT path generated.

\begin{figure}
    \centering
    \includegraphics[scale=0.15]{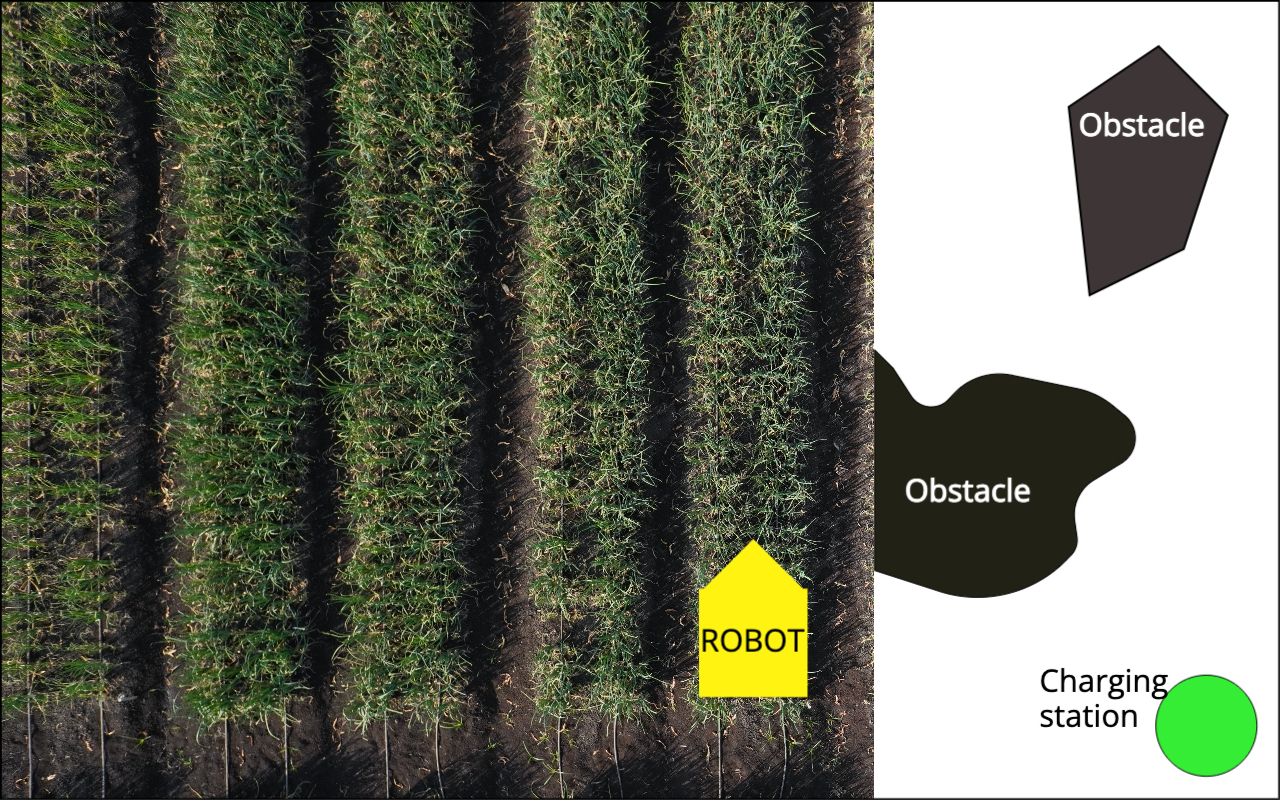}
    \caption{Schematic of an onion field with onion harvester robot}
    \label{fig:problem}
\end{figure}

We assume the robot knows the energy required to execute the task and to travel back to the charging station. An application can be an onion harvester robot, as shown in Fig. \ref{fig:problem}. The robot must harvest along the rows of onions moving from one row to another. It also finds a path to the closest charging station at each instant. Since the robot knows its current battery level and can estimate the energy required for harvesting and traveling back to the charging station, it can decide when to return. 

We survey the literature in the next subsection, followed by the problem formulation in Section \ref{problem formulation}. An overview of RRT and RRT* is presented in Section \ref{overview}. The proposed algorithm is explained in Section \ref{dynamic}, and energy management is addressed in Section \ref{energy}. Simulation results are presented in Section \ref{sim} followed by the concluding remarks in Section \ref{conclusion}.

\subsection{Literature Survey} \label{survey}

RRT and RRT* algorithms are studied extensively in the literature. Several extensions are presented.  These algorithms are sampling-based methods for robotic path planning which are probabilistically complete. RRT generates a path while RRT* gives an optimal path (in the limit number of nodes tend to infinity) based on some cost function. Some of the advantages of RRT and its variants include their applicability in complicated workspace space or configuration space, its capability to include robot motion constraints, and so on. An extensive survey on the extensions of RRT is available in \cite{c15} and the references therein. We present the papers relevant to the work presented in this paper.

In \cite{c1}, the RRT* algorithm was used for replanning in a dynamic environment with random, unpredictable moving obstacles. When an obstacle moves to a node location that was included in the path, the path is replanned around the obstacle using the node that is immediately after the node closest to the obstacle. Authors in \cite{c2} and \cite{c11} also used a similar replanning approach to planning a path around obstacles. Additionally, the work in \cite{c2} also limited the number of nodes by removing childless nodes when the number of nodes exceeded a limit which will reduce its complexity. 

Paper \cite{c3} proposed the Dynamic Rate Random Forest or the DRRT. It, too, discards nodes affected by an obstacle and reforms the tree. The Extended Rate Random Tree algorithm (ERRT) mentioned in \cite{c4} proposed using waypoints that are nodes from the tree generated in past iterations. When building the tree either a random node is selected or a node from the waypoint array is selected or the goal node is selected based on a certain probability distribution. Some further improvements to this were made in \cite{c12} which used the waypoint cache method and BG-RRT algorithm.


\cite{c6} developed a combination of RRT and $\text{A}^*$ in which multiple random nodes and the corresponding nearest nodes are selected, but only the node that minimizes a certain premade cost function is further examined. It also experimented with using other types of norms apart from the Euclidean norm while finding distance metrics. \cite{c14} combined the artificial potential field method and RRT by using two trees that advance towards each other with respect to an attractive or repulsive potential.

If $\text{RRT}^*$ takes too long to converge to a path, some other methods can be used to remove redundant turns and optimize the path, like the ant colony optimization algorithm that was used in \cite{c7}. \cite{c13} developed a new approach to $\text{RRT}^*$, which involves first generating a path from an initial point to a goal and then optimizing it by interconnecting directly visible points and doing intelligent sampling. The robot used in this study can not take sharp turns instantaneously and has limits on maximum and minimum turning angles. These nonholonomic constraints need to be accounted for in the path planning algorithm. Some ways to incorporate these constraints have been discussed in \cite{c9} and \cite{c10}.

\emph{Contributions} - The problems addressed in the literature considered a dynamic environment and different methods of replanning using RRT are presented. We consider the case where the robot is following a pre-defined path, and the RRT* is used to plan a path to the base for re-charging. Therefore the starting point of the RRT* is changing at every instance. We propose a fast algorithm called "dynamic path RRT" that can find the cost to reach the base so that the performance of the robot to carry out its assigned task is not hampered. Hence, our problem setting is different from what exists in the literature. We also compared the performance of our proposed algorithm with some of the algorithms in the literature.

\section{Problem Formulation} \label{problem formulation}

We consider an autonomous robot executing some task in a known environment. There are one or more charging stations. The robot follows a predefined path until it requires recharging its batteries. The robot plans an online path to the charging station using RRT*. We assume the power needed to move at a constant speed is known, and the robot moves at a fixed desired speed. So, the robot knows the energy it will need to return to the charging point. The robot can also measure the energy remaining in the batteries.

To model the robot, we choose a classic car model as
\begin{align}
    \frac{d\mathbf{x}}{dt} &= \mathbf{v} \\
    \frac{d\psi}{dt} &= \omega = \frac{v}{L} \tan \delta 
\end{align}
where $\mathbf{x}$ represents position in the $xy$ plane, $\mathbf{v}$ represents the velocity in the $xy$ plane, $v = \| \mathbf{v} \|$ is its speed, $\psi$ represents the heading and $\omega$ represents the angular velocity of the robot, $L$ is the length of the robot and $\delta$ is its steering angle. We assume $v$ is fixed, so the input to the robot is $\omega$.

\begin{figure}
\centering
\begin{tikzpicture}
  \draw[dashed] (0,0) circle (1);
    \draw (0,0) -- (30:1) arc (30:150:1) -- cycle;
    \draw (-0.866, 0.5) circle (0.08);
    \draw (0.866, 0.5) circle (0.08);
    \draw[->] (-0.866,0.5) -- (-0.491,1.149);
    \draw[dashed] (-0.866, 0.5) -- (0.866, 0.5);
    \draw (-0.766, 0.673)  .. controls (-0.55,0.53) and (-0.67, 0.6) .. (-0.6, 0.5);
    \draw (0,0) -- (30:0.15) arc (30:150:0.15) -- cycle;
    \node at (-0.45, 0.6) {$\alpha$};
    \node at (-0.45, 1.3) {$\mathbf{v}$};
    \node[align=right] at (-2 ,0.5) {Initial point};
    \node at (2 ,0.5) {Final point};
    \node at (-0.5, 0.1) {$R$};
    \node at (0, 0.27) {$\theta$};
    \node at (0.1, 0.7) {$d$};
\end{tikzpicture}
\caption{Trajectory of the robot from one point to another}
\label{fig:omega}
\end{figure}
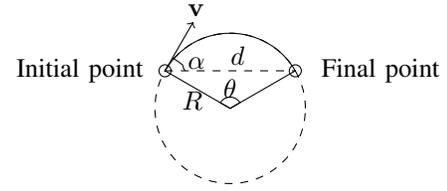

We use simple geometry to make the robot go from current to new positions. Consider Fig. \ref{fig:omega}. Here, $\theta = 2 \alpha$ and $R = \frac{d}{2 \sin \alpha}$. The linear and angular velocities are related by
\begin{align}
    \omega=\frac{v}{R}=\frac{2v\sin\alpha}{d}\label{eqn:omega}
\end{align}
where both $\alpha$ and $d$ are measurable. Since there exist limits on the steering angle $\delta \leq \delta_{\text{max}}$, $\omega$ will be restricted to 
\begin{align}
    \omega \leq \left| \frac{v}{L} \tan (\delta_{\text{max}}) \right|    \label{eqn:omega_limit}
\end{align}
So, if \eqref{eqn:omega} demands an $\omega$ outside the above limits, the final point will not be reachable. We associate a cost with the path from the current state to the next state. The cost is equal to the energy required to travel the path. For simplicity, we assume that the energy spent is proportional to the distance covered. However, formulations can also be used. Therefore, the cost ($E$) of the path is
\begin{align}
    E=k R\theta=2kR\alpha\label{eqn:cost}
\end{align}
where $k$ is the gain relating distance to cost.


\section{Overview of RRT and $\text{RRT}^*$} \label{overview}
RRT  is a well-known probabilistic algorithm used for path planning in robotics. $\text{RRT}^*$ is a modification of RRT that can generate an optimal path. We give an overview of these algorithms to relate to our proposed algorithm and for the completeness of the paper. A pseudocode for the RRT algorithm is given below.
\begin{algorithm}
\caption{RRT algorithm}\label{alg:rrt}
\begin{algorithmic}
\Procedure{RRT}{$q_i, q_g$}
    \State $T \gets$ initialize\_tree($q_i$, $q_g)$
    \While{goal\_not\_reached}
        \State $q_{\text{rand}} \gets$random\_sample()
        \State neighbourhood, $q_{\text{nearest}} \gets$get\_nearest\_node($q_{\text{rand}}$)
        \State $q_{\text{new}} \gets$ steer($q_{\text{nearest}}$, $q_{\text{random}}$) 
        \If{is\_obstacle\_free($q_{\text{nearest}}$, $q_{\text{new}}$)}
            \State cost($q_{\text{new}}$) = cost($q_{\text{nearest}}$) + energy($q_{\text{nearest}}$, $q_{\text{new}}$) 
            \State T$\gets$ insert\_node($q_{\text{nearest}}$, $q_{\text{new}}$)
            \If{$q_{\text{new}} = q_g$}
                \State goal\_not\_reached = False
            \EndIf
        \EndIf
    \EndWhile
\EndProcedure
\end{algorithmic}
\end{algorithm}
First, the initial and final nodes are initialized i.e. $q_i$ and $q_g$. A random node $q_{\text{rand}}$ is selected from the workspace. Then by checking all the nodes in the tree, the node closest to the random node is found: $q_{\text{nearest}}$. Also, the random node is moved towards the nearest node up to a certain step size. Then in the steer() function angular velocity to go from the nearest node to the random node is computed, and if it lies outside the bounds set by turning rate constraints, it is set to $\text{sign}(\omega) \omega_{\text{max}}$. The coordinates of the new node can then be fixed by moving from the nearest node with the constant $\omega$ that is calculated. The minimum distance is the RRT step size and the maximum distance is the radius of the ball used for finding the neighborhood. The pseudocode for this is given in algorithm 1. If the path from the nearest node to the new node intersects an obstacle then a different random node is picked. In this work we consider obstacles to be line segments thus, it is easy to check if the path is intersecting the obstacle. Before adding the new node to the tree its cost is computed by adding the cost of its parent and the energy required to go from the nearest node to the new node. The new node now becomes a child node of $q_{\text{nearest}}$. If this new node happens to be the goal then RRT is ended and a path is generated.

There is a very low probability of the goal node being selected while drawing random samples. Thus instead, the algorithm is stopped when a node becomes sufficiently close to the goal node. Another alternative \cite{c4} is to modify random\_sample() such that a completely random node is drawn with a probability $1 - p_{\text{goal}}$ and the goal node is drawn with probability $p_{\text{goal}}$
\begin{equation}
    P(q) =
\begin{cases}
1 - p_{\text{goal}}, & \text{if } q \text{ is a random node}\\
p_{\text{goal}}, & \text{if } q \text{ is the goal node}
\end{cases}
\end{equation}
RRT algorithm will converge on a path from the initial point to the goal. However, the path may or may not be the optimal path. $\text{RRT}^*$ modifies this code to find the optimal path with two changes. First, in the steer function, after the random node is brought within a certain step size of the nearest node, all the nodes lying in a neighborhood around the random node of a size greater than the step size are selected and put into an array. Thus the nearest node will always be in this neighborhood. Instead of taking the nearest node as a parent, the node with the least cost in that neighborhood is chosen as a parent. Secondly, the nodes are "rewired" that is after a new node has been added it is checked if it can be connected with a neighboring node to reduce the cost of that node.


\section{Dynamic Replanning} \label{dynamic}


We assume that the map of the environment is available to the robot. For example, the onion-harvesting robot can map the field while harvesting since the path to the base will be only through the harvested region. A suboptimal path to the goal is generated as the robot moves, and the energy required to transverse this path is computed. This energy is compared with the current battery level of the robot. If sufficient energy is available, the robot continues to harvest.
Else the plucking is stopped, and an optimal path is computed using $\text{RRT}^*$. The energy required for the new path is used to make the decision on return to base. 

We propose a new algorithm that relies on trees built in past iterations. Consider a tree already generated by the algorithm, and a path is found. We call this the old path and the tree being built in the current iteration is called new tree. Now the initial node is placed in the new position of the robot. As new nodes are added to the new tree, it checks if there are any nodes from the old path nearby. If so, the new tree is built further by replicating nodes from the old path.

\begin{algorithm}
\caption{Dynamic Path RRT Algorithm}\label{alg:rrt}
\begin{algorithmic}
\Procedure{Dynamic\_Path\_RRT}{$q\_i, q\_g$}
    \State $T \gets$ initialize\_tree($q_i$, $q_g$)
    \State $F \gets$ initialize\_forest($T$)
    \While{goal\_not\_reached}
        \State $q_{\text{rand}} \gets$random\_sample()
        \State $q_{\text{nearest}} \gets$get\_nearest\_node($q_{\text{rand}}$)
        \State $q_{\text{new}} \gets$ steer($q_{\text{nearest}}$, $q_{\text{random}}$)
        \If{is\_obstacle\_free($q_{\text{nearest}}$, $q_{\text{new}}$)}
            \State cost($q_{\text{new}}$) = cost($q_{\text{nearest}}$) + energy($q_{\text{nearest}}$, $q_{\text{new}}$) 
            \State T$\gets$ insert\_node($q_{\text{nearest}}$, $q_{\text{new}}$)
            \State p $\gets$ UniformRandom(0, 1)
            \If{p $<$ $\text{p}_{\text{scan}}$}
                \State old\_nodes $\gets$ scan\_forest(F, $q_{\text{new}}$)
                \State check\_connection(old\_nodes, $q_{\text{new}}$)
            \EndIf
            \If{goal\_reached(T)}
                \State goal\_not\_reached = False
            \EndIf
        \EndIf
    \EndWhile
\EndProcedure
\end{algorithmic}
\end{algorithm}

Here it is important to introduce a new data structure - forest. It is the collection of the paths generated in past iterations. Not all paths need to be remembered since this might lead to excessive memory requirements. We propose to discard the oldest path from the forest when a new path is added. The scan\_forest function checks if any of the nodes from the older iterations are present in a neighbourhood around the new node.


The check connection function iterates through all the nodes gathered from scan forest function and finds the node that can be attached to the newly added node with minimum cost. It then calls the path building function. The pseudocode for this is in algorithm 3.


\begin{algorithm}
    \caption{Check Connection and Build Path}\label{alg:rrt}
    \begin{algorithmic}
        \Procedure{check\_connection}{old\_nodes, F, $q_{\text{new}}$}
            \State candidate\_nodes[] $\gets$ initialize\_array
            \For{$q_{\text{old path}}$ in old\_nodes}
                    \State $q_{\text{old path}} \gets$ steer($q_{\text{new}}$, $q_{\text{old path}}$)
                    \State cost($q_{\text{old path}}$) = cost($q_{\text{new}}$) + energy($q_{\text{new}}$, $q_{\text{old path}}$) 
                    \State candidate\_nodes.add($q_{\text{old path}}$)
            \EndFor
        \State $q_{\text{old path}}$ = $\min_{\text{cost}}$candidate\_nodes
        \State path\_building($q_{\text{new}}$, $q_{\text{old path}}$)
        \EndProcedure
    \end{algorithmic}
\end{algorithm}

The path\_building function takes in $q_{\text{new}}$ i.e. the latest node added to the new tree and $q_{\text{old path}}$ that is the minimum cost node returned by 'check connection'. Path building attaches $q_{\text{old path}}$ to the new tree and then finds its child node which is also part of the old path. It then attaches that child node to the new tree and this goes on in a recursive process. The pseudocode for this is given in algorithm 4.

\begin{algorithm}
    \caption{Path Building from Older Paths}\label{alg:rrt}
    \begin{algorithmic}
        \Procedure{path\_building}{$q_{\text{new tree}}$, $q_{\text{old path}}$}
            \State new\_tree $\gets$ insert\_node($q_{\text{new tree}}$, $q_{\text{old path}}$)
            \For{child\_node in $q_{\text{old path}}$.children}
                \If{child\_node.part\_of\_old\_path is True}
                    \State new\_child\_node $\gets$ steer($q_{\text{old path}}$, child\_node)
                \EndIf
                \State path\_building($q_{\text{old tree}}$, new\_child\_node)
            \EndFor
        \EndProcedure
    \end{algorithmic}
\end{algorithm}



\subsection{Robot Energy Management} \label{energy}
Now that the dynamic path RRT algorithm is ready the procedure to determine robot's decision to go to the recharge point needs to be fixed. 

\begin{algorithm}
    \caption{Energy management algorithm}\label{alg:rrt}
    \begin{algorithmic}
        \Procedure{decide\_return}{$\mathbf{x}$, F}
            \State latest\_path $\gets$ dynamic\_path\_RRT($\mathbf{x}$, $q_g$)
            \If{latest\_path.cost $>$ safety\_factor $\times$ $\mathbf{x}.\text{power}$}
                \State latest\_path $\gets \text{RRT}^*(\mathbf{x},q_g, \mathbf{x}.\text{power})$
                    \If{latest\_path.cost $>$ safety\_factor $\times$ $\mathbf{x}.\text{power}$}
                        \State latest\_path$\gets$execute($\mathbf{x}$)
                \EndIf
            \EndIf
            \State $\mathbf{x}$ += robot\_velocity
        \EndProcedure
    \end{algorithmic}
\end{algorithm}

The pseudocode for this is given in algorithm 5. Here $\mathbf{x}$ denotes the state of the robot i.e. its position, velocity, angle, and charge. Dynamic path RRT is first used to generate a suboptimal path. Then it is checked if the current battery energy is enough to execute that path with a pre-determined safety factor ($< 1$). If yes, then the robot moves to its new position. If not it runs RRT* until a path with low enough cost is found or until a maximum node limit is reached. If RRT* was unsuccessful the robot returns to the charging station.

\section{Simulation Results} \label{sim}
We simulated the algorithm in Python on a computer with an Intel CORE i7 processor and Ubuntu 22.04 OS. Two different environments were considered. We compared the execution time of our algorithm with that of ERRT[4] and regular RRT. In the simulations, we assumed the forward speed of the robot to be 1 unit and a maximum steering angle of $40^\circ$. The probability of scanning for old nodes was 0.7 and the probability for the sampling goal was 0.2. We used a step size of 0.05 unit for RRT and a waypoint sampling probability of 0.7 for ERRT. Rewiring was done in the first iteration when an initial tree was generated but not done later.

\begin{figure}
    \centering
    \includegraphics[width = 0.5\textwidth]{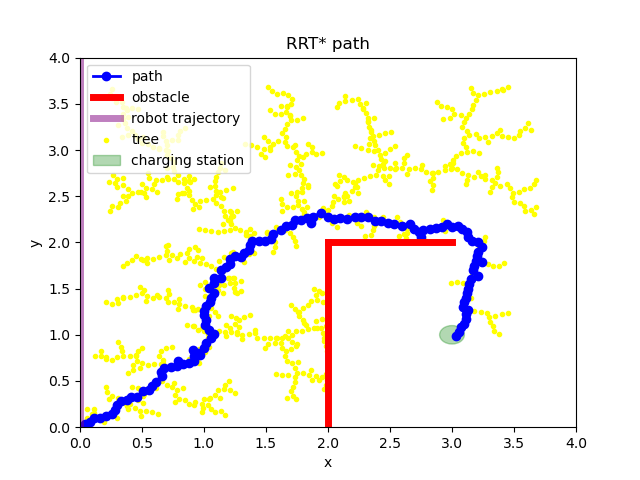}
    \caption{First tree formed using pure $\text{RRT}^*$}
    \label{fig:sim1a}
\end{figure}

\begin{figure}
    \centering
    \includegraphics[width = 0.5\textwidth]{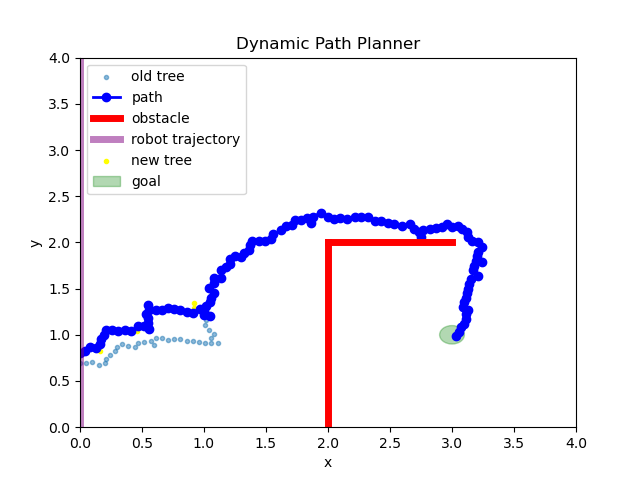}
    \caption{Decrease in number of nodes after using the dynamic path planner}
    \label{fig:sim1b}
\end{figure}

\begin{figure}
    \centering
    \includegraphics[width = 0.5\textwidth]{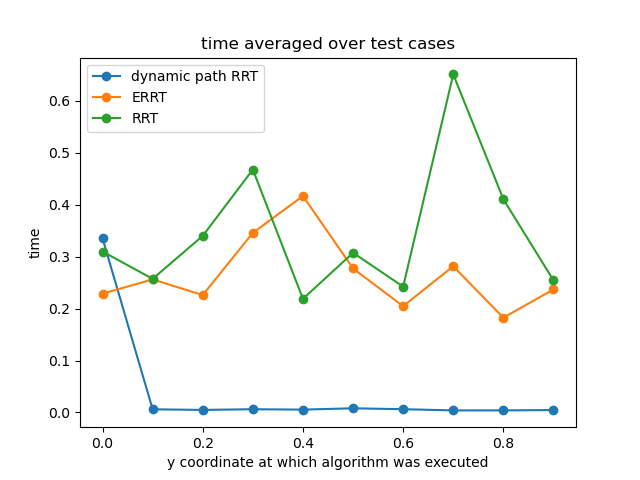}
    \caption{Average time taken for different y coordinates}
    \label{fig:avgtime}
\end{figure}

In the first scenario, the robot moves parallel to the $y$-coordinate. The snapshot of the path generated at the initial time and some intermediate time is shown in Figs. \ref{fig:sim1a}-\ref{fig:sim1b}. The dynamic path RRT algorithm performed much better than ERRT and regular RRT. Please refer to Fig. \ref{fig:avgtime}. It is worth mentioning that in the time analysis of dynamic path RRT, the very first dot in the plot at $y = 0$ represents pure RRT*. That is why the time taken for the first iteration is much higher than all others. Planning a trajectory by using information from older iterations is far more efficient than using pure RRT in each case. 

\begin{figure}
    \centering
    \includegraphics[width = 0.5\textwidth]{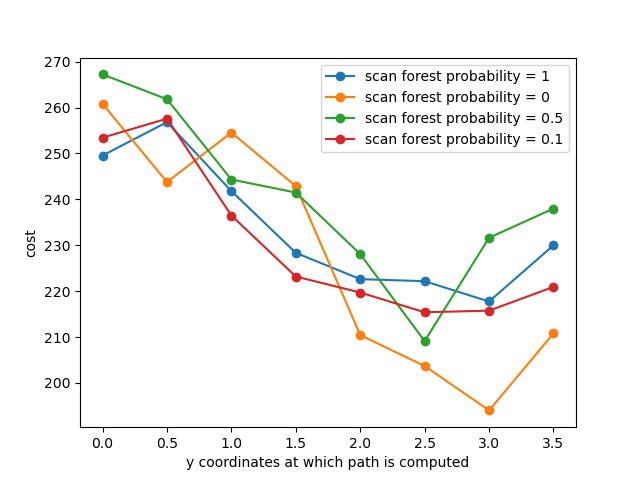}
    \caption{Average cost of paths vs y coordinate}
    \label{fig:avgcost}
\end{figure}

Figure \ref{fig:avgcost} shows how the cost of each path varies as the position of the robot changes. This was done for different values of scan forest probability. This shows initially, the cost of RRT* is similar to the cost of the dynamic path planner. The costs of the path generated by varying $p_{\text{scan}}$ in the algorithm are also similar. However, as the robot moves farther away, the algorithm gives paths with higher costs. Thus as the robot moves further away from its starting point, running an RRT* intermittently will help.

\begin{figure*}
    \centering
    \includegraphics[width = 0.45\textwidth]{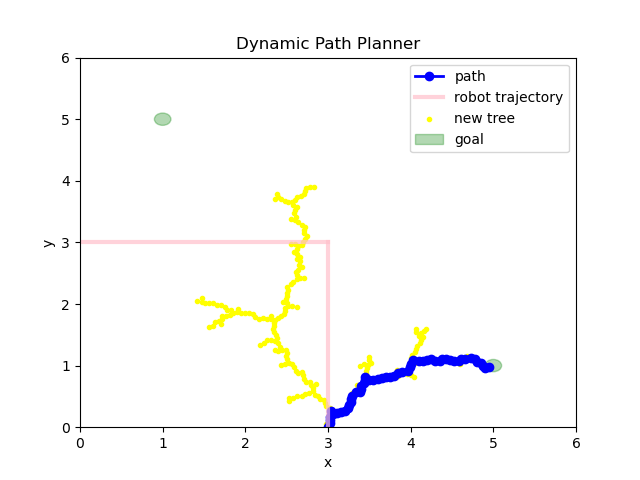}
    \includegraphics[width = 0.45\textwidth]{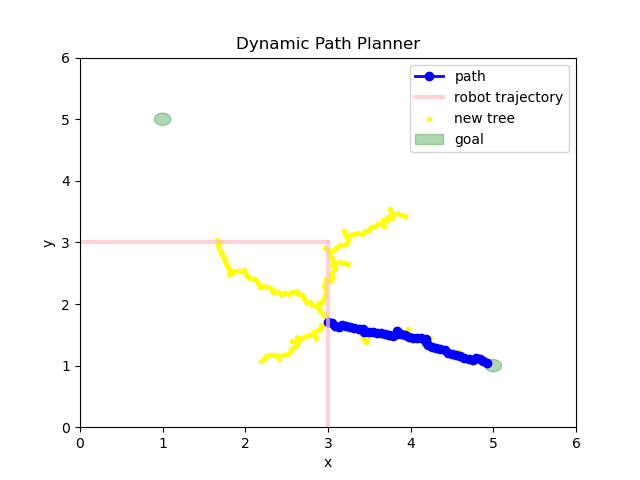}
    \includegraphics[width = 0.45\textwidth]{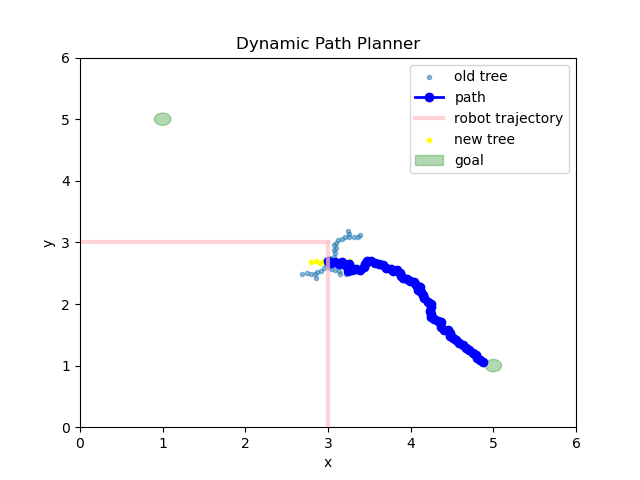}
    \includegraphics[width = 0.45\textwidth]{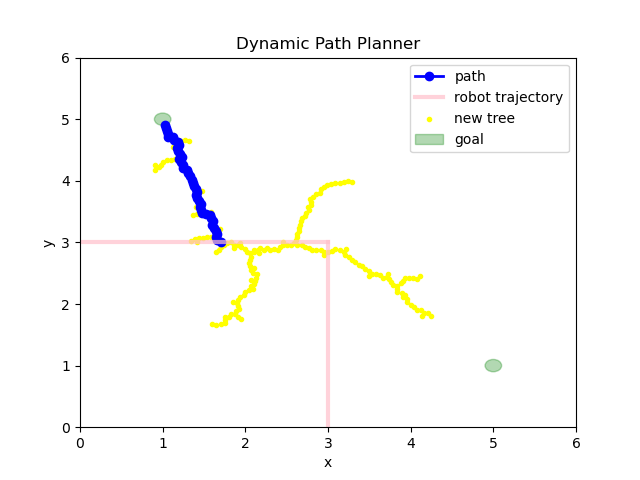}
    \caption{Top-Left: Initial path made by $\text{RRT}^*$; Top-Right: The path made by the robot can not be used with its current battery thus it makes a new path using $\text{RRT}^*$; Bottom-Left: New path made using the dynamic path RRT; Bottom-Right: The same situation happens here too but this time it finds a path to a different recharging station. It now considers this as its goal point.}
    \label{fig:sim2}
\end{figure*}

The energy management algorithm was also tested on a field with dimensions $6 \times 6$ and no obstacles. The robot goes from $(3, 0)$ to $(3, 3)$ and then takes a 90-degree turn and starts moving towards $(0, 3)$. There are two recharging stations at $(1, 5)$ and $(5, 1)$ respectively. The snapshots at different instances are shown in Fig. \ref{fig:sim2}.

\section{Conclusion} \label{conclusion}
This paper presents an innovative approach that addressed the critical challenges of path planning and energy management in autonomous mobile robots. The dynamic path RRT algorithm is proposed that generates paths from the robot's current position to the recharge station efficiently by using information from old iterations. It is then checked if the robot's current battery is sufficient to safely execute this path and a decision is taken based on this condition. 
We compared the proposed algorithm with ERRT as well as regular RRT and found that the proposed strategy is significantly faster. However, the cost of the path is usually more than RRT*. We also analyzed the proposed algorithm by varying the probability of building the connection with the older paths. It is observed that the new paths are found faster as the probability is increased, but the path cost also increases. A user can decide on the probability based on the trade-off between time and cost. 



\begin{thebibliography}{99}

\bibitem{c1}
Devin Connell and Hung Manh La, "Dynamic Path Planning and Replanning
for Mobile Robots using RRT*", 2017 IEEE {\it International Conference on Systems, Man, and Cybernetics (SMC),} Banff, AB, Canada, 2017, pp. 1429-1434, doi: 10.1109/SMC.2017.8122814.


\bibitem{c2}
 Adiyatov and H. A. Varol, "A novel RRT*-based algorithm for motion planning in Dynamic environments," {\it 2017 IEEE International Conference on Mechatronics and Automation (ICMA), Takamatsu, Japan, 2017,} pp. 1416-1421, doi: 10.1109/ICMA.2017.8016024.

\bibitem{c3}
D. Ferguson, N. Kalra and A. Stentz, "Replanning with RRTs," { \it Proceedings 2006 IEEE International Conference on Robotics and Automation, 2006. ICRA 2006., Orlando, FL, USA, 2006,} pp. 1243-1248, doi: 10.1109/ROBOT.2006.1641879.

\bibitem{c4}
Bruce, James and Manuela M. Veloso. “Real-time randomized path planning for robot navigation.” { \it IEEE/RSJ International Conference on Intelligent Robots and Systems 3} (2002): 2383-2388 vol.3.


\bibitem{c6}
J. Li, S. Liu, B. Zhang and X. Zhao, "RRT-A* Motion planning algorithm for non-holonomic mobile robot," { \it 2014 Proceedings of the SICE Annual Conference (SICE), Sapporo, Japan,} 2014, pp. 1833-1838, doi: 10.1109/SICE.2014.6935304.

\bibitem{c7}
J. Qi, H. Yang and H. Sun, "MOD-RRT*: A Sampling-Based Algorithm for Robot Path Planning in Dynamic Environment," in { \it IEEE Transactions on Industrial Electronics,} vol. 68, no. 8, pp. 7244-7251, Aug. 2021, doi: 10.1109/TIE.2020.2998740.


\bibitem{c9}
Khanal, Abhishek. “RRT and RRT* Using Vehicle Dynamics.” { \it ArXiv} abs/2206.10533 (2022): n. pag.

\bibitem{c10}
Heß, Robin. (2013). Trajectory Planning for Car-Like Robots Using Rapidly Exploring Random Trees. { \it IFAC Proceedings Volume}s. 46. 44-49. 10.3182/20131111-3-KR-2043.00018. 

\bibitem{c11}
E. Shan, B. Dai, J. Song and Z. Sun, "A Dynamic RRT Path Planning Algorithm Based on B-Spline," { \it 2009 Second International Symposium on Computational Intelligence and Design}, Changsha, China, 2009, pp. 25-29, doi: 10.1109/ISCID.2009.155.

\bibitem{c12}
Chengren Yuan, Guifeng Liu, Wenqun Zhang, Xinglong Pan,
An efficient RRT cache method in dynamic environments for path planning,
{ \it Robotics and Autonomous Systems,
Volume 131,}
2020,
103595,
ISSN 0921-8890,

\bibitem{c13}
F. Islam, J. Nasir, U. Malik, Y. Ayaz and O. Hasan, "RRT$\ast$-Smart: Rapid convergence implementation of RRT$\ast$ towards optimal solution," { \it 2012 IEEE International Conference on Mechatronics and Automation, Chengdu, China, 2012}, pp. 1651-1656, doi: 10.1109/ICMA.2012.6284384.

\bibitem{c14}
W. Xinyu, L. Xiaojuan, G. Yong, S. Jiadong and W. Rui, "Bidirectional Potential Guided RRT* for Motion Planning," in { \it IEEE Access}, vol. 7, pp. 95046-95057, 2019, doi: 10.1109/ACCESS.2019.2928846.

\bibitem{c15}
Iram Noreen, Amna Khan and Zulfiqar Habib, “Optimal Path Planning using RRT* based Approaches: A Survey and Future Directions” { \it International Journal of Advanced Computer Science and Applications(IJACSA)}, 7(11), 2016. http://dx.doi.org/10.14569/IJACSA.2016.071114



\end{thebibliography}
\end{document}